\documentclass{article}
\usepackage{spconf,amsmath,graphicx}
\usepackage[utf8]{inputenc}
\usepackage{subcaption}
\usepackage{hyperref}
\hypersetup{colorlinks,citecolor=blue,urlcolor=blue,linkcolor=blue}
\usepackage{tabularx}
\usepackage{verbatim}

\title{Few-shot learning with attention-based sequence-to-sequence models}
\name{Bertrand Higy$^{1, 2}$\sthanks{This work was performed while Bertrand Higy was at the Centre for Speech Technology Research} and Peter Bell$^3$}
\address{$^1$ iCub Facility, Istituto Italiano di Tecnologia, Italy\\
$^2$ Università degli studi di Genova, Italy\\
$^3$ Centre for Speech Technology Research, University of Edinburgh, UK}

\begin{document}
\ninept
\maketitle
%
%\vspace{-10pt}
\begin{abstract}
End-to-end approaches have recently become popular as a means of simplifying the training and deployment of speech recognition systems. However, they often require large amounts of data to perform well on large vocabulary tasks. 
%To make speech recognition accessible to a wider audience, 
With the aim of making end-to-end approaches usable by a broader range of researchers, we explore the potential to use end-to-end methods in small vocabulary contexts where smaller datasets may be used.  A significant drawback of small-vocabulary systems is the difficulty of expanding the vocabulary beyond the original training samples -- therefore we also study strategies to extend the vocabulary with only few examples per new class (few-shot learning).

Our results show that an attention-based encoder-decoder can be competitive against a strong baseline on a small vocabulary keyword classification task, reaching 97.5\% of accuracy on Tensorflow's Speech Commands dataset. It also shows promising results on the few-shot learning problem where a simple strategy achieved 68.8\% of accuracy on new keywords with only 10 examples for each new class. This score goes up to 88.4\% with a larger set of 100 examples.
\end{abstract}
\vspace{-5pt}
\section{Introduction}
\label{sec:intro}
\vspace{-3pt}

Vocal interfaces are becoming more and more popular as our devices (e.g. smartphones, tablets or more recently smart speakers) are becoming more intelligent. Speech is an intuitive and effective way to transmit commands, which makes it very appealing. However, the complexity of modern speech recognition technology and the difficulty of gathering the necessary data can make it hard for single individuals or small companies to develop their own systems, even for small vocabulary recognition. This paper explores strategies to train a keyword/command recognition system by trading off the size of the vocabulary against the quantity of data available. In other words, we focus on low resource, small vocabulary tasks, with a strong bias toward simplicity.  Our motivation is similar to that behind the
Google Tensorflow team's release of the Speech Commands (SC) dataset \cite{warden_speech_2018} and the organization of an accompanying challenge\footnote{https://www.kaggle.com/c/tensorflow-speech-recognition-challenge}

Together with the SC dataset, Google released a baseline classification system\footnote{https://www.tensorflow.org/tutorials/sequences/audio\_recognition} which was used as a starting point by many challenge participants. To enable a simple classification system to be used directly without the use of time-warping or other dynamic programming algorithms, every input file in the dataset is constrained to a fixed length, something that would not be required by
the more flexible standard HMM-based approaches to speech recognition. Although the fixed-length constraint is not unreasonable
for a small vocabulary keyword recognition task, we were motivated to consider a more recent end-to-end (E2E) approach -- specifically the attention-based encoder-decoder architecture -- as a means of allowing input of arbitrary length, whilst retaining the simplicity of
a single DNN-based discriminative classifier.  This approach also allows us to readily switch between sub-word (phoneme or grapheme) and word modeling by just changing the target output.

\begin{comment}  % PJB: reworded this section above
This system has some limitations though. To simplify the input process, the requirement is imposed that every input file has a fixed length. Also, the use of a classification system imposes to model keywords or commands as a whole. Those are probably reasonable constraints for a (very) small vocabulary keyword recognition task but will become inadequate in some other contexts where longer records are needed or where sub-word or word based sequential outputs may be more effective. Standard approaches to speech recognition are more flexible, in this respect, but are very complex. This motivated us to consider the more recent E2E approaches, and more specifically the attention-based encoder-decoder architecture. Despite them usually lagging behind hidden Markov model (HMM) based systems in term of performance, they are much simpler to train. They also allow to easily switch between sub-word (phoneme or grapheme) and word modeling by just changing the target output.
\end{comment}

In this paper, we experiment with the use of a sequence-to-sequence (S2S) model for a modified version of the Speech Commands task, comparing it with the traditional deep neural network (DNN)-HMM approach. In the literature, S2S models are usually applied on large vocabulary tasks with large datasets and it is not obvious that they will work well in our setup. 

The obvious limitation of the small vocabulary approach we take is that trained system is confined to the list of commands defined in the original data. To alleviate this constraint, we also explore strategies to extend the set of commands requiring very few examples (the few-shot learning problem \cite{ravi_optimization_2017, snell_prototypical_2017, yang_learning_2018}).

The remainder of the paper is organized as follows: relevant literature is presented in section \ref{sec:rel_work}, methodology and experiments in sections \ref{sec:met} and \ref{sec:res} respectively, and we conclude in section \ref{sec:cl}.

\vspace{-8pt}
\section{Related work}
\label{sec:rel_work}
\vspace{-4pt}

E2E training has attracted much attention recently. One of the first breakthroughs came from the connectionist temporal classification (CTC) loss \cite{graves_connectionist_2006}, which allows an acoustic neural model to be trained directly on unsegmented data. While the original technique is not E2E, it has later been extended to train models that predict grapheme sequences \cite{graves_towards_2014} or in conjunction with a language model (LM) based on recurrent neural networks (RNNs), an architecture refered to as the RNN-transducer \cite{graves_sequence_2012}. More recently, the attention-based encoder-decoder model has been applied to automatic speech recognition (ASR) (see e.g. \cite{chorowski_end--end_2014, chan_listen_2016}).

If the simplicity of the training procedure of E2E systems is attractive, they generally show reduced performance over traditional HMM-based systems, especially so when used without an external LM, a good example being \cite{chan_listen_2016}. Using a much bigger dataset \cite{prabhavalkar_comparison_2017} managed to reach competitive results on a dictation task, but was still performing worse on voice-search data. This doesn't mean though that E2E models will necessarily be bad in lower resource conditions. For example, \cite{rosenberg_end--end_2017} achieved competitive results on several languages, even though it failed to surpass a DNN-HMM baseline. To the best of our knowledge, E2E models have never been applied to small vocabulary speech recognition tasks before. The work closest to ours is probably \cite{shan_attention-based_2018} where an attention-based E2E architecture is applied to keyword spotting. Though, despite the vocabulary being reduced to one word, a very large dataset is used.

Between the different E2E approaches, the attention-based encoder-decoder architecture has been shown to give better results \cite{prabhavalkar_comparison_2017}. While the original model \cite{bahdanau_neural_2015} was proposed for machine translation, several ways to adapt it for speech recognition have since been proposed. A first difference with machine translation resides in the ratio between the length of the input and output sequences: in speech recognition, the input sequence tends to be much longer than the output sequence. \cite{chan_listen_2016} proposed to use pyramidal layers to downsample the input. This reduces the number of hidden states the attention has to attend to, thus improving both the accuracy and the computational performance. Similarly, convolutional neural networks (CNNs) have been shown to be effective \cite{hori_advances_2017}, leading to further improvement. Another concern pertains to the global attention mechanism which is a bit too flexible for speech recognition (an essentially monotonic left-to-right process). Ways to encourage monotonicity \cite{chorowski_end--end_2014} or ensure the local and monotonous nature of the attention system \cite{tjandra_local_2017} have thus been proposed. Taking a different approach, a hybrid CTC/Attention architecture trained in a multi-task fashion has been proposed in \cite{kim_joint_2017, watanabe_hybrid_2017}. The idea there is to use the monotonous and left-to-right properties of CTC to find better alignments, which compensate for the over-flexibility of the attention-based decoder.

\vspace{-7pt}
\section{Methodology}
\label{sec:met}
\vspace{-3pt}

The main task considered here corresponds to the one proposed in Tensorflow's Speech Commands challenge mentioned earlier, that is keyword classification. In addition to the keywords, two additional classes are considered: (i) a \texttt{\_silence\_} category corresponding to records free of speech, and (ii) an \texttt{\_unknown\_} category for records containing speech that is none of the keywords.

%We will now present the CNN-HMM baseline system and the E2E architecture we used for this task. We then detail the strategies %proposed for few-shot learning.

\vspace{-5pt}
\subsection{CNN-HMM baseline}
\label{sec:met_baseline} 

In previous experiments on the SC task, we obtained our best performance with a DNN-HMM system using CNNs. The CNN-HMM model has been trained using a standard Kaldi recipe\footnote{egs/rm/s5/local/nnet/run\_cnn2d.sh} \cite{povey_kaldi_2011}. It is composed of two 2D (across both time and frequency) convolutional layers followed by 4 fully connected layers. The network is trained with cross-entropy, followed by 1 iteration of discriminative training with the state-level minimum Bayes risk (sMBR) objective \cite{gibson_hypothesis_2006}. We use as input 11 frames (5 from both sides) of 40 filterbank coefficients, augmented with $\Delta$ and $\Delta\Delta$ features. 

\vspace{-5pt}
\subsection{End-to-end model}
\label{sec:met_end2end}

We opted for the attention-based encoder-decoder approach, and more precisely the hybrid CTC/Attention model from \cite{kim_joint_2017, watanabe_hybrid_2017} which showed promising results and for which the code was available\footnote{https://github.com/espnet/espnet}. We used a CNN-based encoder that is composed of 4 convolutional layers, with 2 max pooling layers (after the second and fourth convolutions). Each pooling layer has a reduction factor of 2, thus downsampling the timescale of the input by 4 overall. Four layers of 320 bidirectional long-short term memory (BiLSTM) units sit on top of the CNN part.

We used the location-aware attention mechanism \cite{chorowski_attention-based_2015} and a layer of 300 LSTM cells for the decoder. Default hyperparameters from the \texttt{voxforge} recipe were used unless stated otherwise. The input was composed of 80 fbanks and we experimented with 10 different types of labels: phonemes, graphemes and words.

\vspace{-5pt}
\subsection{Strategies for few-shot learning}
\label{sec:met_fewshot}

The main limitation of our small vocabulary approach is its flexibility. The ASR system is limited to the set of keywords it was trained on and no guarantee is given that it will generalize to new ones (in fact we expect it to recognize them poorly if at all). This is a limitation that is hardly manageable in practical usage. To alleviate it, we propose to explore strategies for few-shot learning, where one can gather few examples of a new word and use them to retrain or adapt the existing system, so that it will perform better on this new word.

The simplest strategy we tried consists in adding the examples of the new keywords to the training set from the beginning and train a new model on it (a method referred to as \texttt{retrain} hereafter). One issue with this method is that the new keywords will be under-represented compared to the original ones. To improve on that, we propose to try oversampling the few-shot examples, that is we will see these examples $k$ times during an epoch (where $k$ is the oversampling factor) when the original examples will be seen only once.

The main limitation of the \texttt{retrain} strategy is that it requires to retrain the model from scratch every time. Alternatively, we propose a method based on adaptation (referred to as \texttt{adapt}) where we start from a model trained on the 12 original categories (see section \ref{sec:met_end2end}). We then adapt all its weights by training it for a few more epochs on the few-shot examples, keeping the same training procedure otherwise. To avoid performance deteriorating on the original keywords, we also include some of their examples, with the same number of examples per class as for the few-shot classes. The overall number of examples being very small, few updates are made per epoch. We thus expect higher learning rates to be useful. We also optimized the number of epochs which plays a complementary role.

One drawback of this strategy, however, is that the model we start from may not contain all the output labels required for the new keywords (limited to the phonemes or graphemes present in the 10 original keywords). We solve this problem by replacing the missing phonemes (resp. graphemes) by the \texttt{UNK} model (resp. the character \texttt{\_}) initially introduced for the \texttt{\_unknown\_} category (see section \ref{sec:met_dataset}). This can result in a dramatic change as exemplified by the word \texttt{backward} for which most phonemes are absent from the pretrained models output. The original transcription "\texttt{B AE K W ER D}" becomes "\texttt{UNK UNK UNK UNK UNK D}" after replacement. Similarly for graphemes, replacing missing characters leads to the transcription "\texttt{????w?rd}". In view of this limitation and in order to compare the \texttt{adapt} strategy more fairly with the \texttt{retrain} approach, we introduced the \texttt{retrain\_replace} strategy. This strategy uses the same training procedure as the \texttt{retrain} method, but with the modified labels.

\vspace{-7pt}
\section{Experiments}
\label{sec:res}

\vspace{-4pt}
\subsection{Experimental setup}
\label{sec:met_dataset}

Our experimental setup is derived from the second version of the SC dataset \cite{warden_speech_2018}. It contains 105,800 recordings of 35 different keywords. Each record has a fixed duration of 1 second. The results in \cite{warden_speech_2018} correspond to the task that was proposed for the original challenge, that is the classification of 10 keywords (out of the 35 available), the remaining ones being used to populate the \texttt{\_unknown\_} category.

This setup has been slightly modified here. A limitation of the original design is that the same collection of words are used for the \texttt{\_unknown\_} category at both training and test time. In order to better evaluate the generalization capability of the model to unseen words (which will inevitably happen given the number of words we use to train this category), we decided to exclude some of them from the training set. Also, as mentioned earlier, we are interested in exploring strategies for few-shot learning. Hence, we also kept a few words aside for use in those experiments.

\begin{table}
\begin{tabularx}{\linewidth}{ l X }
    Set & Words \\
    \hline
    \texttt{org\_kwd} & down, go, left, no, off, on, right, stop, up, yes \\
    \texttt{org\_unk} & bed, bird, cat, dog, happy, house, marvin, sheila, tree, visual, wow \\
    \texttt{new\_kwd} & forward, four, one, three, two, zero \\
    \texttt{new\_unk} & eight, five, follow, learn, nine, seven, six \\
\end{tabularx}
\caption{List of keywords assigned to the different sets.}
\label{tab:word_sets}
\vspace{-5pt}
\end{table}

The 10 original keywords (from here on referred to as the \texttt{org\_kwd} set of words) are kept identical. The 25 remaining ones however are split into two main categories: 7 are used as new keywords in the few-shot experiments (refered to as \texttt{new\_kwd}) and 18 as unknowns (the \texttt{unk} set). This later group is further split into 11 words (\texttt{org\_unk}) that are used for training and evaluation, while the remaining 7 (\texttt{new\_unk}) are seen at evaluation time only. When evaluating the \texttt{unk} set, the average of the scores of the two subsets is used (with equal weight for the two categories). Table \ref{tab:word_sets} gives the list of words assigned to each category.

The split of the data in training, validation and test sets was done using the procedure provided in Tensorflow's example code with 80\%, 10\% and 10\% for each set respectively. The \texttt{\_unknown\_} category being the combination of several keywords, it is over-represented in the dataset. To prevent it from dominating the learning procedure, we downsample it for the training set, randomly selecting a number of examples corresponding to the mean number of examples we have for the keywords (\texttt{org\_kwd}). Finally, for all experiments on few-shot learning, we randomly sample $f$ examples of each new class from the training records we have kept aside.

In the phoneme-based experiments, we introduce a special phoneme, labelled \texttt{UNK}, to model the words of the \texttt{\_unknown\_} category. Similarly, for grapheme-based or word-based experiments, all the words are transcribed with the unique character \texttt{?}. In all conditions, we further map all output not corresponding to one of the keywords or \texttt{\_silence\_} to the \texttt{\_unknown\_} category. 

\vspace{-5pt}
\subsection{End-to-end approach for small vocabulary ASR}
\label{sec:res_smallvoc}

\begin{table}
\newcolumntype{C}{>{\centering\arraybackslash}X}%
\begin{tabularx}{\linewidth}{ l C C C }
    Model & \texttt{org\_kwd} & \texttt{unk} & \texttt{new\_kwd} \\
    \hline
    Phoneme-based S2S & \textbf{3.7} & \textbf{22.2} & - \\
    Grapheme-based S2S & 3.8 & 23.1 & - \\
    Word-based S2S & 3.8 & 25.7 & - \\
    \hline
    retrain-10 & \textbf{4.3} & 28.6 & 57.0 \\
    retrain\_replace-10 & 4.7 & \textbf{27.3} & 57.9 \\
    adapt-10 & 11.5 & 65.8 & \textbf{29.2} \\
    \hline
    retrain-100 & \textbf{4.5} & \textbf{29.6} & 16.9 \\
    retrain\_replace-100 & 4.8 & 30.3 & 17.0 \\
    adapt-100 & 7.4 & 42.5 & \textbf{10.1} \\
\end{tabularx}
\caption{Validation classification error (\%) of the S2S model with different types of outputs (first set of rows, trained on 12 classes only) or with different few-shot learning strategies (second and third set of rows, trained with 7 additional classes).}
\label{tab:res_s2s}
\vspace{-12pt}
\end{table}

We first report results on the original classification task trained on 12 categories, comparing traditional and E2E pipelines. Table \ref{tab:res_base} summarizes the results obtained with the S2S model for different types of outputs, where we see that classification error of the three S2S models on the keywords are very close. Looking at the performance of the same models on the \texttt{unk} set, we see they all show a big increase in classification error, as expected with only 11 different words to populate the \texttt{\_unknown\_} category for training. It can be noticed though that the phoneme-based S2S model generalizes better than its two competitors.

Hence, the greater simplicity of grapheme- or word-based approaches, which do not require a pronunciation dictionary, can be traded off for better performance. Moreover, it has to be highlighted that in a small vocabulary context, building the pronunciation dictionary is greatly simplified compared to a large vocabulary context.

\begin{table}
\newcolumntype{C}{>{\centering\arraybackslash}X}%
\begin{tabularx}{\linewidth}{ l C C C }
    Model & \texttt{org\_kwd} & \texttt{unk} & \texttt{new\_kwd} \\
    \hline
    CNN-HMM (baseline) & 4.2 & 35.6 & - \\
    Phoneme-based S2S & \textbf{2.5} & \textbf{23.6} & - \\
    \hline
    retrain-10 & \textbf{3.7} & \textbf{31.3} & \textbf{59.5} \\
    adapt-10 & 11.7 & 66.5 & 31.2 \\
    \hline
    retrain-100 & \textbf{3.8} & \textbf{31.8} & 18.5 \\
    adapt-100 & 6.9 & 45.0 & \textbf{11.6}
\end{tabularx}
\caption{Test classification error (\%) of the baseline and the main S2S models. The first set of rows correspond to models trained on 12 categories only, while second and third set of rows correspond to few-shot experiments.}
\label{tab:res_base}
\vspace{-10pt}
\end{table}

In table \ref{tab:res_base}, we compare the test classification error of the best E2E model with the CNN-HMM baseline. On the main task, the E2E approach beats the baseline by 40\% relative. This is very promising as it shows that E2E models are a competitive alternative to more traditional approaches for our task. The results on the \texttt{unk} set shows that they also generalize much better, the E2E approach beating the baseline by 34\% absolute on this subset.

\begin{figure*}[t!]
    \centering
    \begin{subfigure}[b]{0.40\textwidth}
        \includegraphics[width=\textwidth]{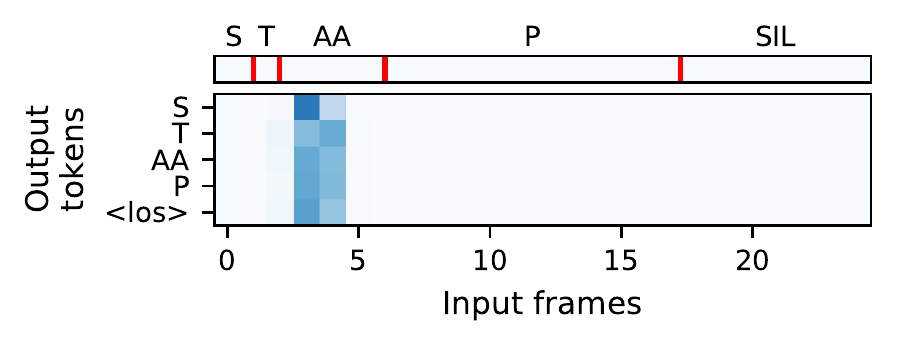}
        \includegraphics[width=\textwidth]{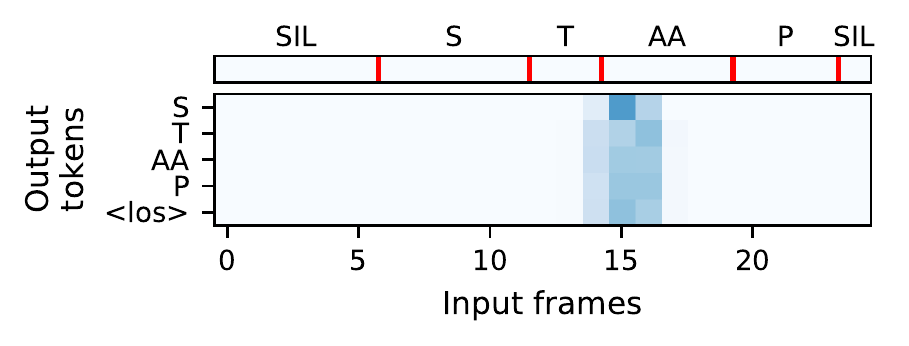}
    \end{subfigure}
    \begin{subfigure}[b]{0.40\textwidth}
        \raisebox{2.3mm}{\includegraphics[width=\textwidth]{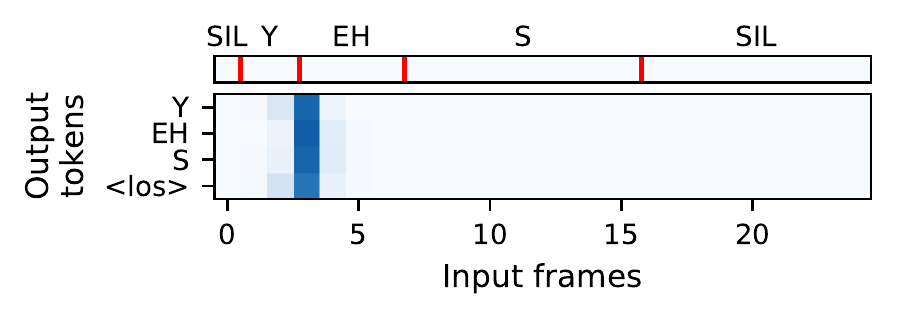}}
        \raisebox{1.4mm}{\includegraphics[width=\textwidth]{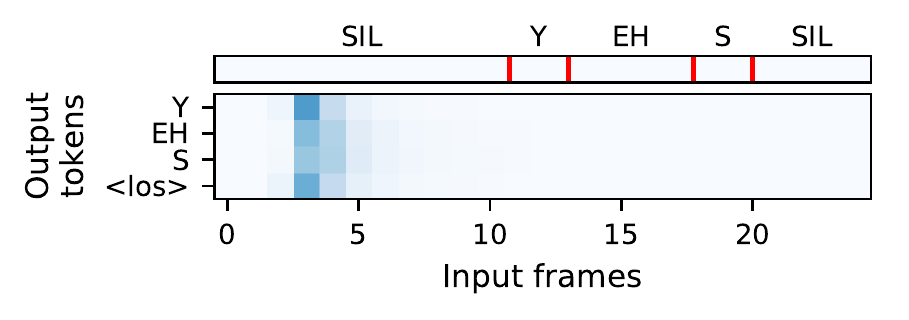}}
    \end{subfigure}
    \begin{subfigure}[b]{0.06\textwidth}
        \raisebox{10.5mm}{\includegraphics[width=\textwidth]{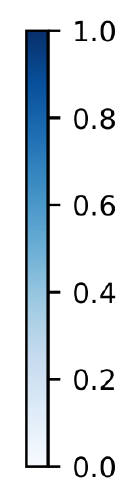}}
    \end{subfigure}
    \vspace{-5pt}
    \caption{Attention weights produced by the phoneme-based S2S model for two examples of the words "stop" (left) and "yes" (right), with their respective alignment on top. }
    \label{fig:att}
    \vspace{-15pt}
\end{figure*}

Finally, we give some insight on the behavior of the S2S models. As can be seen from figure \ref{fig:att}, and confirmed by manual inspection, the attention tends to focus on a single portion of each input and doesn't shift as the output tokens are produced. It appears that the model representation is more akin to word than sub-word modeling, as is usually observed. With our small vocabulary, the model is apparently able to discriminate between the different keywords with a single "glance" at the data. For example, in the case of the word \texttt{stop}, the model seems to attend to the phoneme \texttt{AA} (left part of figure \ref{fig:att}). More surprisingly, in the case of the word \texttt{yes}, the model seems to seek information from a fixed position, even if it falls in the silence preceding the word. A more quantitative analysis would be required to better understand those dynamics, which maybe related to the effective window size of the encoder.

\vspace{-5pt}
\subsection{Few-shot learning}
\label{sec:res_fewshot}

For low values of $f$, the variability introduced by the random selection of the few-shot examples is high. We thus report here the mean classification error over 10 runs for all experiments on few-shot learning. 

We experiment with $f \in \{10, 100\}$. While 100 examples may seem a lot for few-shot learning, it allows to test how the different strategies behave when the number of examples increase. It is also to be noted that 100 examples is only 2.6\% of the number of examples available for the original classes ($\sim$ 3850 on average).

\begin{figure}[t!]
    \includegraphics[width=0.40\textwidth]{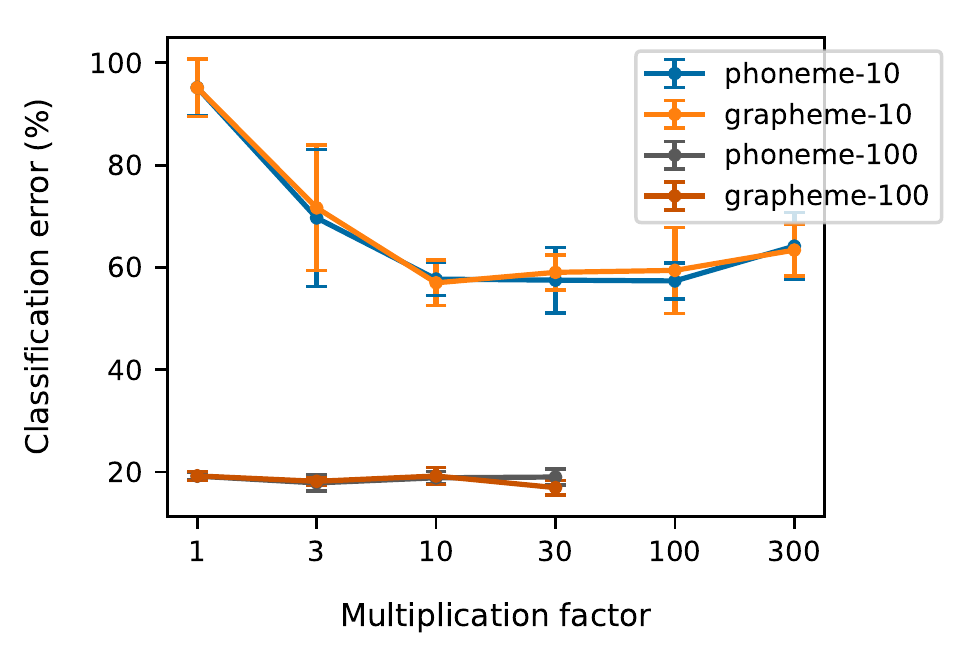}
    \vspace{-5pt}
    \caption{Validation classification error for the new keywords with the \texttt{retrain} strategy. We compare phoneme- and grapheme-based outputs, for $f$ = 10 or 100. The bars represent the standard deviation over 10 runs.}
    \label{fig:scratch}
    \vspace{-10pt}
\end{figure}

For the \texttt{retrain} strategy, we tried oversampling the new keywords up to 3000 simulated examples ($k = 300$ for $f = 10$ and $k = 30$ for $f = 100$), so as to reach similar frequency in training and test sets. As figure \ref{fig:scratch} shows, the \texttt{retrain} strategy gives poor results on the \texttt{new\_unk} set without oversampling ($k$ = 1). In contrast, higher values of $k$ give much better scores. The best results are obtained with $k = 10$ for $f = 10$, where the grapheme-based model reaches 57.0\% of classification error. Conversely with $f = 100$, best results are obtained for $k = 30$ with the grapheme-based model, for a classification error of 16.9\%. 

\begin{figure}[t!]
    \includegraphics[width=0.48\textwidth]{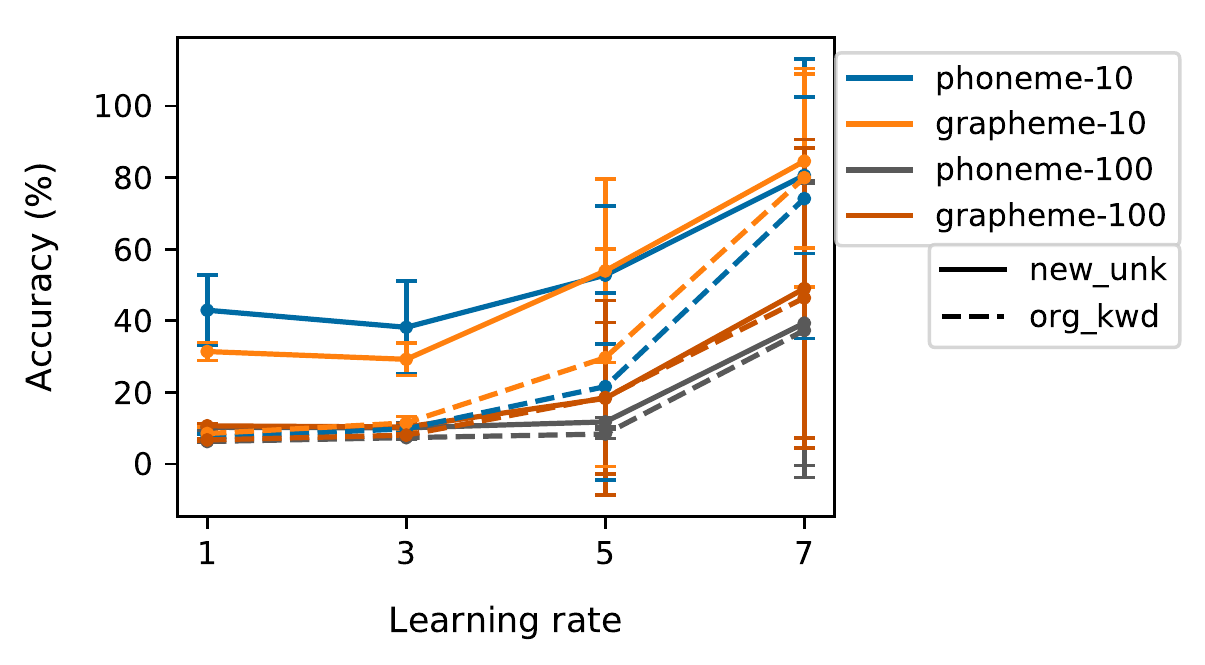}
    \vspace{-15pt}
    \caption{Validation classification error for the new keywords (plain) and the original keywords (dashed) with the \texttt{adapt} strategy, as a function of the learning rate. Phoneme- and grapheme-based outputs are compared for 10 and 100 fewshot-examples ($f$).}
    \label{fig:adapt}
    \vspace{-10pt}
\end{figure}

Figure \ref{fig:adapt} shows the classification error with the \texttt{adapt} strategy on the \texttt{new\_kwd} set as a function of the learning rate ($lr$). For each experiment, the best number of epochs is selected based on the validation set. We see that increasing the learning rate is necessary to get good result, but too high a value and the results deteriorate as we overfit the small training set. The best results are achieved with a learning rate of 3, using the grapheme-based (resp. phoneme-base) output for $f = 10$ (resp. $f = 100$).  We also reported the classification error on the \texttt{org\_kwd} set to show how the original training is progressively undone as the learning rate increases. As one can see, the performance degrades consistently and at some point drops suddenly as the model overfits the adaptation set.

Table \ref{tab:res_s2s} summarizes the classification error of the different strategies on the validation set for the two values of $f$ (10 and 100) with the hyperparameters giving optimal scores on the \texttt{new\_kwd} set. A first and surprising observation is that the phoneme/grapheme replacement rules introduced in section \ref{sec:met_fewshot} for the \texttt{adapt} strategy doesn't seem to penalize the performance on the \texttt{new\_kwd} set. The \texttt{retrain\_replace} strategy gives results very close to the \texttt{retrain} one overall. Comparing the performance of the \texttt{adapt} and \texttt{retrain} strategies now, we see that adaption is not only much faster to train but is also the most performant on the new keywords. Though, this is is achieved at the expense of the \texttt{org\_kwd} and \texttt{unk} sets. Table \ref{tab:res_base} summarizes the test classification error of the best models for both strategies (\texttt{retrain} and \texttt{adapt}).

\vspace{-7pt}
\section{Conclusion}
\label{sec:cl}
\vspace{-3pt}

In this paper, we proposed to study the adequacy of E2E approaches on a small vocabulary task, in order to simplify the process of training a keyword/command recognition system and make this technology more accessible. We found that they can be competitive in such a context, giving better results than a strong CNN-HMM baseline. We also proposed two few-shot strategies. By simply training a model from scratch on the combination of the original dataset and those new examples, we managed to reach 40.5\% of accuracy with only 10 examples per new keyword and 81.5\% with 100 examples. A faster adaptation strategy was also proposed which achieves even better results, reaching 68.8\% and 88.4\% of accuracy for 10 and 100 examples respectively, but at the expense of the performance on the original keywords. This results may be further improved by using more advanced strategies. 

We have also shown that the dynamic of the hybrid CTC/Attention model in our task is quite different from what is usually observed with large vocabulary tasks. It would be interesting in the future to analyze more deeply the behavior of the model as one move from a small vocabulary keyword task to a large vocabulary one with complex sentences.

\vspace{-7pt}
\section{ACKNOWLEDGEMENTS}
\label{sec:ack}
\vspace{-3pt}

We would like to thank Ondřej Klejch, Joachim Fainberg and Joanna Rownicka for their help with the SC Dataset and for sharing their code, on which our baseline is based. We would also like to thank Sameer Bansal for the very fruitful discussions on sequence-to-sequence models for ASR. 

%\vfill\pagebreak

\bibliographystyle{IEEEbib}
\bibliography{refs}

\begin{thebibliography}{10}

\bibitem{warden_speech_2018}
Pete Warden,
\newblock ``Speech {Commands}: {A} {Dataset} for {Limited}-{Vocabulary}
  {Speech} {Recognition},''
\newblock {\em arXiv:1804.03209 [cs]}, Apr. 2018,
\newblock arXiv: 1804.03209.

\bibitem{ravi_optimization_2017}
Sachin Ravi and Hugo Larochelle,
\newblock ``Optimization as a {Model} for {Few}-{Shot} {Learning},''
\newblock in {\em Proc. of the {International} {Conference} on {Learning}
  {Representations} ({ICLR})}, Toulon, France, 2017.

\bibitem{snell_prototypical_2017}
Jake Snell, Kevin Swersky, and Richard Zemel,
\newblock ``Prototypical networks for few-shot learning,''
\newblock in {\em Advances in {Neural} {Information} {Processing} {Systems}},
  Long Beach, CA, USA, 2017, pp. 4077--4087.

\bibitem{yang_learning_2018}
Flood Sung~Yongxin Yang, Li~Zhang, Tao Xiang, Philip~HS Torr, and Timothy~M
  Hospedales,
\newblock ``Learning to compare: {Relation} network for few-shot learning,''
\newblock in {\em Proc. of the {IEEE} {Conference} on {Computer} {Vision} and
  {Pattern} {Recognition} ({CVPR})}, Salt Lake City, UT, USA, 2018.

\bibitem{graves_connectionist_2006}
Alex Graves, Santiago Fernández, Faustino Gomez, and Jürgen Schmidhuber,
\newblock ``Connectionist {Temporal} {Classification}: {Labelling}
  {Unsegmented} {Sequence} {Data} with {Recurrent} {Neural} {Networks},''
\newblock in {\em Proc. of the 23rd {International} {Conference} on {Machine}
  {Learning}}, New York, NY, USA, 2006, {ICML}'06, pp. 369--376, ACM.

\bibitem{graves_towards_2014}
Alex Graves and Navdeep Jaitly,
\newblock ``Towards {End}-to-end {Speech} {Recognition} with {Recurrent}
  {Neural} {Networks},''
\newblock in {\em Proc. of the 31st {International} {Conference} on {Machine}
  {Learning}}, Beijing, China, 2014, vol.~32 of {\em {ICML}'14}, pp.
  II--1764--II--1772.

\bibitem{graves_sequence_2012}
Alex Graves,
\newblock ``Sequence {Transduction} with {Recurrent} {Neural} {Networks},''
\newblock in {\em presented at the {International} {Conference} of {Machine}
  {Learning} ({ICML}) {Workshop} on {Representation} {Learning}}, Edinburgh,
  Scotland, Nov. 2012,
\newblock arXiv: 1211.3711.

\bibitem{chorowski_end--end_2014}
Jan Chorowski, Dzmitry Bahdanau, Kyunghyun Cho, and Yoshua Bengio,
\newblock ``End-to-end {Continuous} {Speech} {Recognition} using
  {Attention}-based {Recurrent} {NN}: {First} {Results},''
\newblock in {\em presented at {NIPS} {Workshop} on {Deep} {Learning} and
  {Representation} {Learning}}, Montréal, Canada, Dec. 2014,
\newblock arXiv: 1412.1602.

\bibitem{chan_listen_2016}
W.~Chan, N.~Jaitly, Q.~Le, and O.~Vinyals,
\newblock ``Listen, attend and spell: {A} neural network for large vocabulary
  conversational speech recognition,''
\newblock in {\em Proc. of the {IEEE} {International} {Conference} on
  {Acoustics}, {Speech} and {Signal} {Processing} ({ICASSP})}, Shanghai, China,
  Mar. 2016, pp. 4960--4964.

\bibitem{prabhavalkar_comparison_2017}
Rohit Prabhavalkar, Kanishka Rao, Tara~N. Sainath, Bo~Li, Leif Johnson, and
  Navdeep Jaitly,
\newblock ``A {Comparison} of {Sequence}-to-{Sequence} {Models} for {Speech}
  {Recognition},''
\newblock in {\em Proc. of {Interspeech}}, Stockholm, Sweden, 2017, pp.
  939--943.

\bibitem{rosenberg_end--end_2017}
A.~Rosenberg, K.~Audhkhasi, A.~Sethy, B.~Ramabhadran, and M.~Picheny,
\newblock ``End-to-end speech recognition and keyword search on low-resource
  languages,''
\newblock in {\em Proc. of the {IEEE} {International} {Conference} on
  {Acoustics}, {Speech} and {Signal} {Processing} ({ICASSP})}, New Orleans, LA,
  USA, Mar. 2017, pp. 5280--5284.

\bibitem{shan_attention-based_2018}
Changhao Shan, Junbo Zhang, Yujun Wang, and Lei Xie,
\newblock ``Attention-based {End}-to-{End} {Models} for {Small}-{Footprint}
  {Keyword} {Spotting},''
\newblock {\em arXiv:1803.10916 [cs, eess]}, Mar. 2018,
\newblock arXiv: 1803.10916.

\bibitem{bahdanau_neural_2015}
Dzmitry Bahdanau, Kyunghyun Cho, and Yoshua Bengio,
\newblock ``Neural {Machine} {Translation} by {Jointly} {Learning} to {Align}
  and {Translate},''
\newblock in {\em Proc. of the {International} {Conference} on {Learning}
  {Representations} ({ICLR})}, San Diego, CA, USA, 2015,
\newblock arXiv: 1409.0473.

\bibitem{hori_advances_2017}
Takaaki Hori, Shinji Watanabe, Yu~Zhang, and William Chan,
\newblock ``Advances in {Joint} {CTC}-{Attention} based {End}-to-{End} {Speech}
  {Recognition} with a {Deep} {CNN} {Encoder} and {RNN}-{LM},''
\newblock in {\em Proc. of {Interspeech}}, Stockholm, Sweden, June 2017,
\newblock arXiv: 1706.02737.

\bibitem{tjandra_local_2017}
Andros Tjandra, Sakriani Sakti, and Satoshi Nakamura,
\newblock ``Local {Monotonic} {Attention} {Mechanism} for {End}-to-{End}
  {Speech} and {Language} {Processing},''
\newblock in {\em Proc. of the 8th {International} {Joint} {Conference} on
  {Natural} {Language} {Processing}}, Taipei, Taiwan, May 2017,
\newblock arXiv: 1705.08091.

\bibitem{kim_joint_2017}
S.~Kim, T.~Hori, and S.~Watanabe,
\newblock ``Joint {CTC}-attention based end-to-end speech recognition using
  multi-task learning,''
\newblock in {\em Proc. of the {IEEE} {International} {Conference} on
  {Acoustics}, {Speech} and {Signal} {Processing} ({ICASSP})}, New Orleans, LA,
  USA, Mar. 2017, pp. 4835--4839.

\bibitem{watanabe_hybrid_2017}
S.~Watanabe, T.~Hori, S.~Kim, J.~R. Hershey, and T.~Hayashi,
\newblock ``Hybrid {CTC}/{Attention} {Architecture} for {End}-to-{End} {Speech}
  {Recognition},''
\newblock {\em IEEE Journal of Selected Topics in Signal Processing}, vol. 11,
  no. 8, pp. 1240--1253, Dec. 2017.

\bibitem{povey_kaldi_2011}
Daniel Povey, Arnab Ghoshal, Gilles Boulianne, Lukas Burget, Ondrej Glembek,
  Nagendra Goel, Mirko Hannemann, Petr Motlicek, Yanmin Qian, Petr Schwarz, and
  {others},
\newblock ``The {Kaldi} speech recognition toolkit,''
\newblock in {\em {IEEE} 2011 workshop on automatic speech recognition and
  understanding}, Waikoloa Village, HI, US, 2011, IEEE Signal Processing
  Society.

\bibitem{gibson_hypothesis_2006}
Matthew Gibson and Thomas Hain,
\newblock ``Hypothesis spaces for minimum bayes risk training in large
  vocabulary speech recognition,''
\newblock in {\em Proc. of the {Ninth} {International} {Conference} on {Spoken}
  {Language} {Processing}}, Pittsburgh, PA, USA, 2006.

\bibitem{chorowski_attention-based_2015}
Jan~K Chorowski, Dzmitry Bahdanau, Dmitriy Serdyuk, Kyunghyun Cho, and Yoshua
  Bengio,
\newblock ``Attention-based models for speech recognition,''
\newblock in {\em Proc. {Advances} in neural information processing systems},
  Montréal, Canada, 2015, pp. 577--585.

\end{thebibliography}

\end{document}